\documentclass[10pt,twocolumn,letterpaper]{article}

\usepackage[pagenumbers]{cvpr} 
\usepackage{algorithmic}
\usepackage{algorithm} 
\usepackage{colortbl}
 \usepackage{multirow}
 \usepackage{adjustbox}

\usepackage[dvipsnames]{xcolor}

\definecolor{cvprblue}{rgb}{0.21,0.49,0.74}
\usepackage[pagebackref,breaklinks,colorlinks,citecolor=cvprblue]{hyperref}

\def\paperID{*****} 
\def\confName{CVPR}
\def\confYear{2024}

\title{Fusion-Mamba for Cross-modality Object Detection}

\author{
Wenhao Dong{$^{1}$}\thanks{These authors contributed equally.}, Haodong Zhu{$^{1}$}\footnotemark[1], Shaohui Lin {$^{2}$}\thanks{Corresponding Author: shaohuilin007@gmail.com.}, Xiaoyan Luo{$^{1}$}, Yunhang Shen{$^3$}, 
\\
Xuhui Liu{$^{1}$}, Juan Zhang{$^{1}$}, Guodong Guo{$^4$},Baochang Zhang{$^{1}$} \\ 
{$^1$}Beihang University, Beijing, China \ {$^2$}East China Normal University, Shanghai, China \\
{$^3$}Tencent Youtu Lab, Shanghai, China \ {$^4$}Eastern Institute of Technology, Ningbo, China
}

\begin{document}

\maketitle
\begin{abstract}
Cross-modality fusing complementary information from different modalities effectively improves object detection performance, making it more useful and robust for a wider range of applications.
Existing fusion strategies combine different types of images or merge different backbone features through elaborated neural network modules.
However, these methods neglect that modality disparities affect cross-modality fusion performance, as different modalities with different camera focal lengths, placements, and angles are hardly fused.
In this paper, we investigate cross-modality fusion by associating cross-modal features in a hidden state space based on an improved Mamba with a gating mechanism.
We design a \textit{Fusion-Mamba block}~(FMB) to map cross-modal features into a hidden state space for interaction, thereby reducing disparities between cross-modal features and enhancing the representation consistency of fused features.
FMB contains two modules: the State Space Channel Swapping~(SSCS) module facilitates shallow feature fusion, and the Dual State Space Fusion~(DSSF) enables deep fusion in a hidden state space.
Through extensive experiments on public datasets, our proposed approach outperforms the state-of-the-art methods on $m$AP with 5.9\% on $M^3FD$ and 4.9\% on FLIR-Aligned datasets, demonstrating superior object detection performance.
To the best of our knowledge, this is the first work to explore the potential of Mamba for cross-modal fusion and establish a new baseline for cross-modality object detection.
\end{abstract}
\begin{figure}[t]
    \centering
    \includegraphics[width=1.0\linewidth]{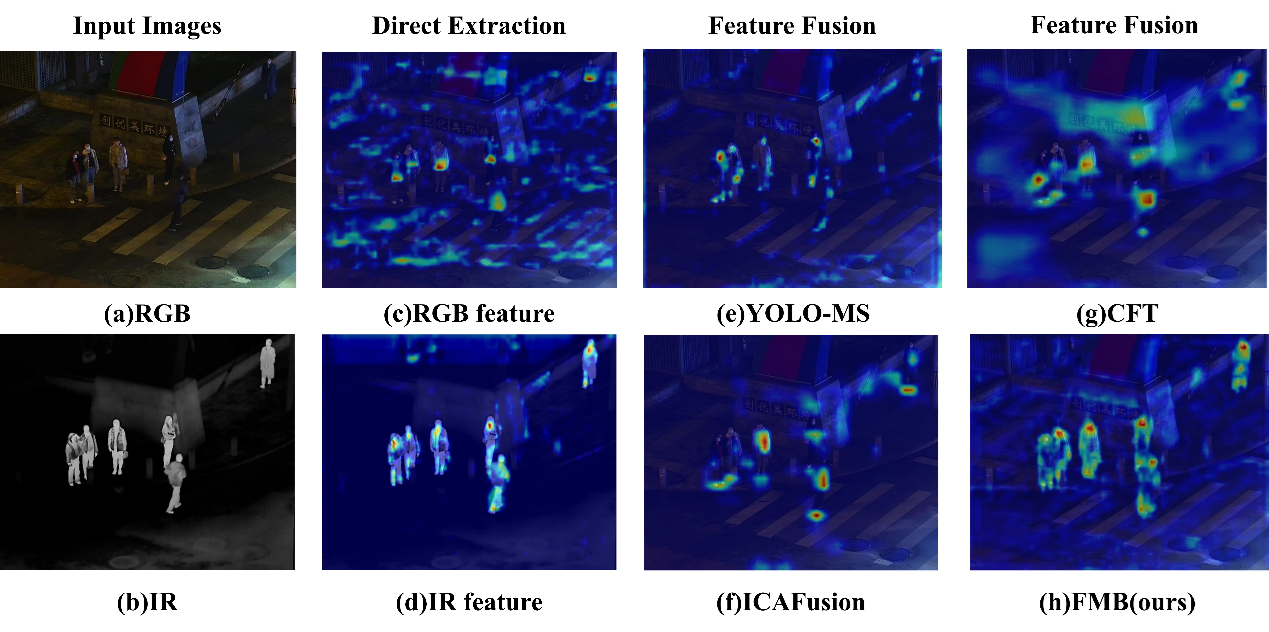}
    \vspace{-1em}
    \caption{
    Heatmap visualization.
   ~(a) and~(b) show the initial RGB and IR input images.
   ~(c) and~(d) show heatmaps generated from single-modality using YOLOv8. 
   ~(e) shows the heatmap of YOLO-MS with a CNN-based fusion module. 
   ~(f) and~(g) show heatmaps of ICAFusion and CFT with a transformer-based fusion module. 
   ~(h) shows the heatmap of our FMB, which achieves better localization.
    }
    \label{fig:1}
\end{figure}

\section{Introduction}
With the swift development of multi-modality sensor technology, multi-modality images have been used in many different areas.
    Among them, paired infrared~(IR) and visible images have been utilized widely, since such two modalities of images provide complementary information. For example,
    infrared images show a clear thermal structure of objects without being affected by luminance, while they lack the texture details of the target.
    In contrast, visible images capture rich object texture and scene information, but lighting conditions severely affect image quality. 
    Thus, many studies focus on infrared and visible feature fusion to improve the perceptibility and robustness for downstream high-level image and scene understanding tasks, \emph{e.g.}, object detection, and image segmentation.

    \begin{figure*}[t]
    \centering
    \includegraphics[width=1\linewidth]{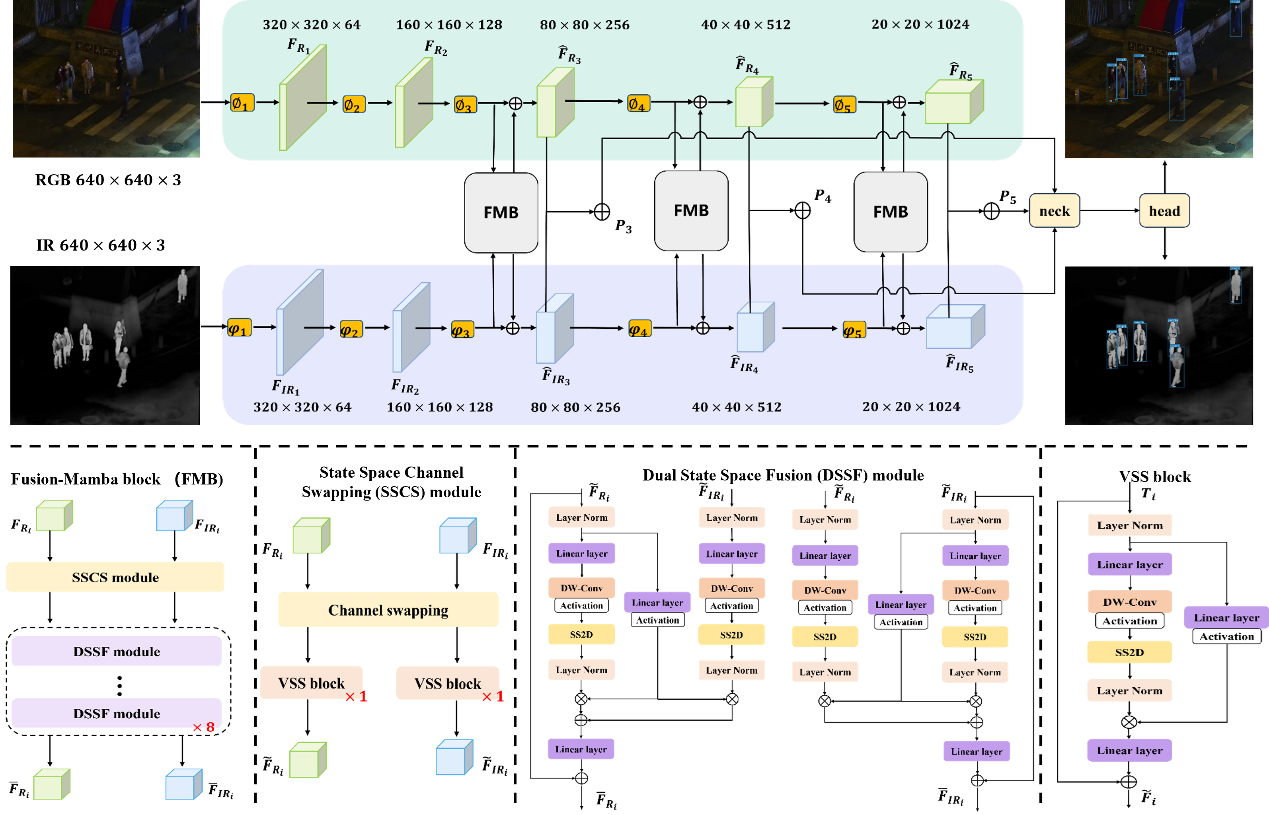}
    \vspace{-1em}
    \caption{
    The architecture of the proposed Fusion-Mamba method. The detection network comprises a dual-stream feature extraction network and three Fusion-Mamba blocks~(FMB), with the same neck and head as YOLOv8. 
    The top is our detection framework, $\phi_i$ and $\varphi_i$ are the convolutional modules of the RGB and IR branches, which are used to generate features of  $F_{R_i}$ and $F_{IR_i}$, respectively.
    $\hat{F}_{R_i}$ and $\hat{F}_{IR_i}$ are the enhanced feature maps through our FMB.
    $P_3, P_4$, and $P_5$ are the summation outputs of enhanced feature maps as the feature pyramid inputs for the neck at the last three stages.
    The bottom shows the design details of our FMB.
    }
    \label{fig:Architecture}
    \end{figure*}
    
    Existing multi-spectral fusion approaches generally employ deep convolutional neural networks~(CNNs)~\cite{simonyan2014very,szegedy2015going,he2016deep} or Transformers~\cite{vaswani2017attention,dosovitskiy2020image} to fuse the cross-modality features.
    %
    A halfway fusion is introduced to integrate two-branch middle-level features from RGB and IR images for multispectral pedestrian detection~\cite{liu2016multispectral}.
    GFD-SSD~\cite{zheng2019gfd} uses Gated Fusion Units to build a two-stream middle fusion detector, which achieves a higher performance than a single modality.
    In this light, YOLO-MS is introduced based on two CNN-based fusion modules to fuse the adjacent branches from the YOLOv5 backbone for real-time object detection~\cite{chen2023yolo}.
    Albeit with great success for cross-modality fusion based on CNNs with local receptive fields, 
    %
    Transformers-based~\cite{vaswani2017attention,dosovitskiy2020image} methods have been proposed to effectively learn long-range dependencies for cross-modality feature fusion.
    CFT~\cite{qingyun2021cross} is the first study to explore a Transformer for middle-level feature fusion, which can improve YOLOv5 performance.   
    ICAFusion~\cite{shen2024icafusion}  with a double cross-attention Transformer can successfully model global features and capture complementary information among modalities. However, these cross-modality fusion methods fail to consider modality disparities, which produces adverse effects on the cross-modality feature fusion. 
    As shown in Fig.~\ref{fig:1}(e)(f)(g), the heatmaps of YOLO-Ms, ICAFusion and CFT fusion features show that they cannot effectively fuse features from different modalities and model the correlations between cross-modality objects, as they have clearly different modality representations.
    This begs our rethinking: \emph{Can we have an effective cross-modality interactive space to reduce modality disparities for a consistent representation, which can thus benefit from the cross-modality relationship for feature enhancement?}
    Moreover, Transformer-based cross-modality fusion is compute-intensive with a quadratic time- and space-complexity.
    
    %

    In this paper,  we propose a Fusion-Mamba method, aiming to fuse features in a hidden state space, which might open up a new paradigm for cross-modality feature fusion.
    We are inspired by Mamba~\cite{gu2023mamba,liu2024vmamba,zhu2024vision}  with a linear complexity to build a hidden state space, which is further improved by a gating mechanism to enable a deeper and more complex fusion. Our Fusion-Mamba method lies in the innovative Fusion-Mamba block~(FMB), as illustrated in Fig.~\ref{fig:Architecture}.
    In FMB, we design a State Space Channel Swapping~(SSCS) module for shallow feature fusion to improve the interaction ability of cross-modality features, and a Dual State Space Fusion~(DSSF) module to build a hidden state space for cross-modality feature association and complementarity. These two blocks help reduce the modal disparities during fusion as shown in Fig.~\ref{fig:1}(h). The heatmap shows that our method fuses features more effectively and makes the detector focus more on the target.  
    This work makes the following contributions:

    1) 
     The proposed Fusion-Mamba method explores the potential of Mamba for cross-modal fusion, which enhances the representation consistency of fused features. We build a hidden state space for cross-modality interaction to reduce disparities between cross-modality features based on an improved  Mamba by gating mechanisms. 
     
     2) 
     We design a Fusion-Mamba block with two modules: the State Space Channel Swapping~(SSCS) module facilitates shallow feature fusion, and the Dual State Space Fusion~(DSSF) module enables deep fusion in a hidden state space.
     
     3) Extensive experiments on three public RGB-IR object detection datasets demonstrate that our method achieves state-of-the-art performance, offering a new baseline in the cross-modal object detection method.

\section{Related Works}
\textbf{Multi-modality Object Detection.} With the rapid development of single modal detectors such as YOLO series models~\cite{redmon2016you}, Transformer~{carion2020end,liu2021swin,guo2022cmt}, multi-modal object detectors have emerged to make good use of images from different modalities. So far, research on multi-modality object detection has focused on two main directions: pixel-level fusion and feature-level fusion.
Pixel-level fusion merges multi-modal input images and the fused image is fed into the detector. Those methods focus on reconstructing fused images using multi-modal input image information~{li2018densefuse, ma2019fusiongan, creswell2018generative}. 
Feature-level fusion joins the output of a detector at a certain stage, such as the early and later features extracted by the backbone~(early and middle fusion~\cite{wang2022improving,cao2023multimodal}) and the detection output~(late fusion~\cite{li2019illumination,chen2022multimodal}). Feature-level fusion can integrate the fusion operation into the detection network as a united end-to-end CNN and or Transformer framework~\cite{cao2023multimodal, wang2022improving, chen2023yolo, qingyun2021cross, shen2024icafusion}. Those fusion methods can effectively improve the object detection performance of single-modality. However, they are still limited in modeling modality disparity and fusion complexity.

\textbf{Mamba.} Since Mamba~\cite{gu2023mamba} was proposed for linear-time sequence modeling in the NLP field, it has been rapidly extended for applications in various computer vision tasks.
Vmamba~\cite{liu2024vmamba} introduces a four-way scanning algorithm based on the characteristics of the image and constructs a Mamba-based vision backbone with better performance in object detection, object segmentation, and object tracking than Swin Transformer.
VM-UNet~\cite{ruan2024vm} is splendid in the field of medical segmentation based on the UNet framework and Mamba blocks.
After that, many Mamba-based deep networks~\cite{ma2024u,wang2024mamba,zhang2024vm} are proposed to make accurate segmentation in medical images.
Video Mamba~\cite{chen2024video} expands the original 2D scan to different bidirectional 3D scans and designs a Mamba framework to use mamba in the video understanding area.

Different from previous methods, our work is the first to exploit Mamba for multi-modality feature fusion. We introduce a carefully designed Mamba-based structure to integrate the cross-modality features in a hidden state space.

\begin{figure*}[t]
    \centering
    \includegraphics[width=1.0\linewidth]{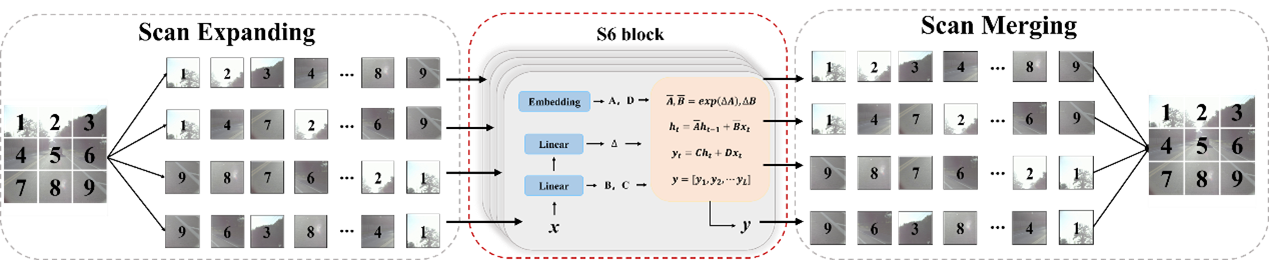}
    \caption{ Illustration of the 2D Selective Scan~(SS2D) on a RGB image. Initially, the image undergoes scan expansion, resulting in four distinct feature sequences. Subsequently, each of these sequences is independently processed through the S6 block. Finally, the outputs of the S6 block are combined through scan merging to generate the final 2D feature map.}
    \label{fig:SS2D}
\end{figure*}

\section{Method}


\subsection{Preliminaries}
\label{sec:Preliminaries}
\textbf{State Space Models.}
State Space Models~(SSMs) are frequently used to represent linear time-invariant systems, which process a one-dimensional input sequence $x(t) \in \mathcal{R}$ by passing it through intermediate implicit states $h(t) \in \mathcal{R}^N$ to produce an output $y(t) \in \mathcal{R}$.
Mathematically, SSMs are often formulated as linear ordinary differential equations~(ODEs): 
\begin{equation}
\begin{aligned}
&h'(t) = Ah(t) + Bx(t),  \\
&y(t) = Ch(t) + Dx(t),
\end{aligned}
\label{eq1}
\end{equation}
where the system's behavior is defined by a set of parameters, including the state transition matrix $A \in \mathcal{R}^{N \times N}$, the projection parameters $B, C \in \mathcal{R}^{N\times 1}$, and the skip connection $D \in \mathcal{R}$.
$Dx(t)$ can be easily removed by setting $D=0$ for exposition.

\textbf{Discretization.}
The continuous-time nature of SSMs in Eq.~\ref{eq1} poses significant challenges when applied in deep learning scenarios.
To address this issue, it is necessary to discretize the ODEs through a process of discretization, which serves the key purpose of converting ODEs into discrete functions. 
It is essential for ensuring alignment between the model and the sampling rate of the underlying signals in the input data, facilitating efficient computational operations~\cite{li2022dense}. 
Considering the input $x_k \in \mathcal{R}^{L \times D}$, a sampled vector within the signal flow of length $L$ following~\cite{zhang2022isnet}, the introduction of a timescale parameter $\Delta$ allows for the transition from continuous parameters $A$ and $B$ to their discrete counterparts $\overline{A}$ and $\overline{B}$, adhering to the zeroth-order hold~(ZOH) principle. Consequently, Eq.~\ref{eq1} is discretized as follows:
\begin{equation}
\begin{aligned}
    &h_k = \overline{A}h_{k-1} + \overline{B}x_k, \\
    &y(t) = \overline{C}h_k + D x_k, \\
    &\overline{A} = e^{\Delta A}, \\
    &\overline{B} =~(\Delta A)^{-1}(e^{\Delta A} - I)\Delta B, \\
    &\overline{C} = C,
\end{aligned}
\label{eq2}
\end{equation}
where $B,C \in \mathcal{R}^D$ and $I$ is an identity matrix. 
After discretization, SSMs are computed by a global convolution with a structured convolutional kernel $\bar{K}\in\mathcal{R}^D$:
\begin{equation}
    y = x*\bar{K}, \quad \bar{K} = \big(C\bar{B}, C\bar{A}\bar{B},\cdots, C\bar{A}^{L-1}\bar{B}\big).
\label{eq3}
\end{equation}
Based on Eq.~\ref{eq2} and Eq.~\ref{eq3}, Mamba~\cite{gu2023mamba} designs a simple selection mechanism to parameterize the SSM parameters of $\Delta$, $A$, $B$, and $C$ depending on the input $x$, which selectively propagates or forgets information along the sequence length dimension for 1D language sequence modeling.

\textbf{2D-Selective-Scan Mechanism.}   
The incompatibility between 2D visual data and 1D language sequences renders the direct application of Mamba to vision tasks inappropriate. For example, while 2D spatial information plays a crucial role in vision-related endeavors, it takes the secondary role in 1D sequence modeling. This discrepancy leads to limited receptive fields that fail to capture potential correlations with unexplored patches. 
2D selective scan~(SS2D) Mechanism is introduced in~\cite{liu2024vmamba} to address the above challenge. The overview of SS2D is depicted in Fig.~\ref{fig:SS2D}. 
SS2D first scan expands image patches into four distinct directions to generate four independent sequences. This quad-directional scanning methodology guarantees that every element within the feature map incorporates information from all other positions across various directions. Consequently, it establishes a comprehensive global receptive field without necessitating a linear increase in computational complexity. Subsequently, each feature sequence undergoes processing using the selective scan space state sequential model~(S6)~\cite{gu2023mamba}. Finally, the feature sequences are aggregated to reconstruct the 2D feature map. SS2D serves as the core element of the Visual State Space~(VSS) block, which is illustrated in Fig.~\ref{fig:Architecture} and will be used to build a hidden state space for cross-modality feature fusion.

\subsection{Fusion-Mamba}
\label{sec:Proposed}
\subsubsection{Architecture}
\label{sec:Architecture}

\begin{figure}[t]
    \centering
    \includegraphics[width=0.63\linewidth]{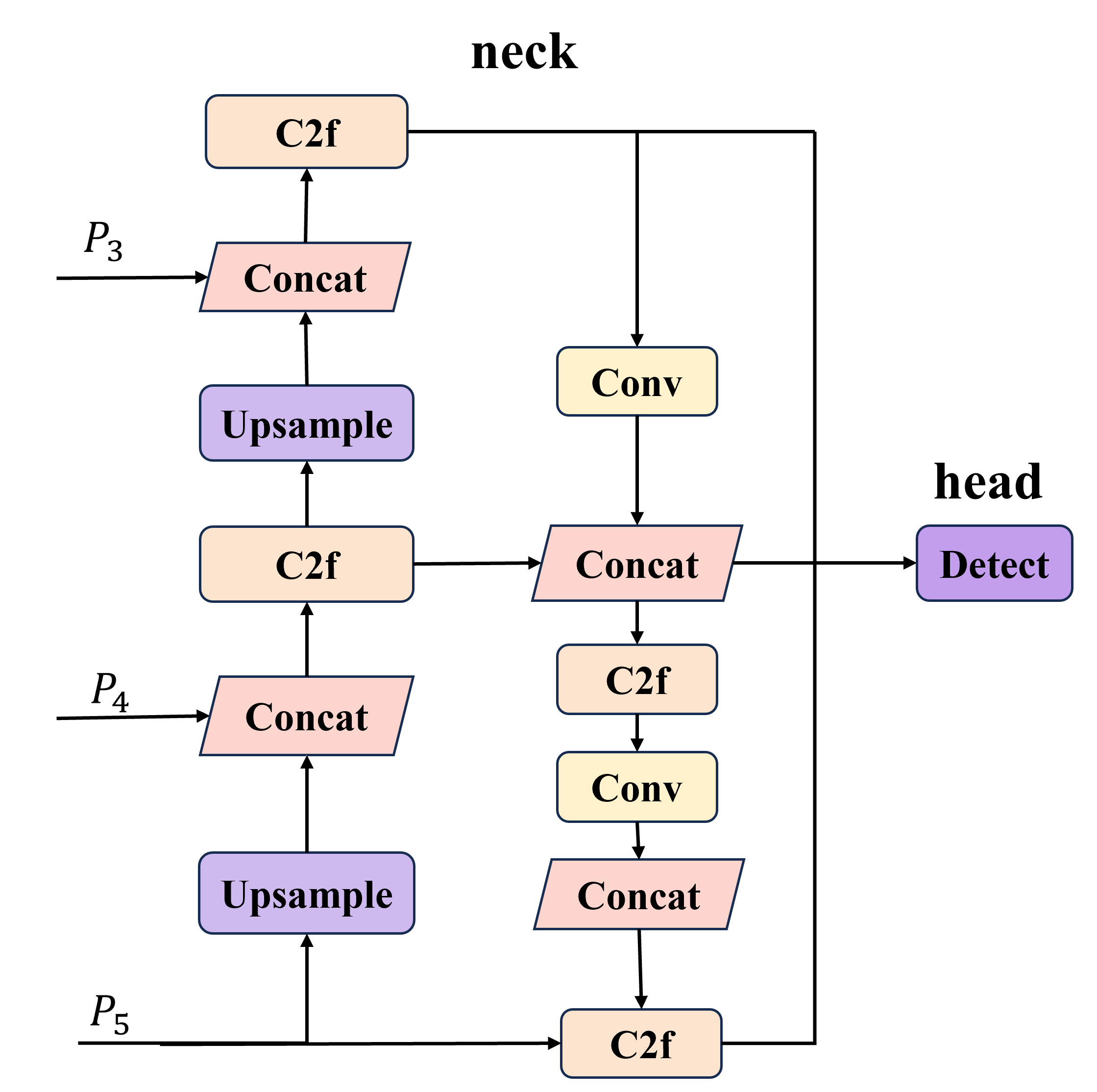}
    \vspace{-1em}
    \caption{Illustration of the neck and head following Yolov8.}
    \label{fig:neck}
\end{figure}

The architecture of our model is depicted in Fig.~\ref{fig:Architecture}. Its detection backbone comprises a dual-stream feature extraction network and three Fusion-Mamba blocks~(FMB), while the detection network contains the neck and head for cross-modality object detection.
The feature extraction network facilitates the extraction of local features from RGB and IR images, denoted by $F_{R_i}$ and $F_{IR_i}$, respectively.
After that, we input these two features into FMB by associating cross-modal features in a hidden state space, which reduces disparities between
cross-modal features and enhances the representation consistency of fused features. 
Specifically, these two local features first undergo the State Space Channel Swapping~(\(SSCS\)) module for shallow feature fusion to obtain interactive features $\tilde{F}_{R_i}$ and $\tilde{F}_{IR_i}$. Then, we feed these interactive features into the Dual State Space Fusion~(\(DSSF\)) module for deep feature fusion in the hidden state space, which generates the corresponding complementary features $\overline{F}_{R_i}$ and $\overline{F}_{IR_i}$. 
The local features are enhanced to generate $\hat{F}_{R_i}$ and $\hat{F}_{IR_i}$ by adding the original ones $F_{R_i}$ and $F_{IR_i}$ into the complementary features $\overline{F}_{R_i}$ and $\overline{F}_{IR_i}$, respectively. 
Subsequencely, the enhanced features $\hat{F}_{R_i}$ and $\hat{F}_{IR_i}$ are directly added to generate the fused feature $P_i$.
In this paper, FMB is only added to the last three stages to generate fused features $P_3, P_4$ and $P_5$~(if not specified), which are the inputs for the neck and head of Yolov8 to generate the final detection results~(As shown in Fig.~\ref{fig:neck}). 

\subsubsection{Key components}
\label{sec:Components}

Given the input RGB image $I_R$ and infrared image $I_{IR}$, we feed them into a series of convolutional blocks to extract their local features:
\begin{equation}
 F_{R_i}=\phi_i{\cdots}(\phi_2(\phi_1(I_R))), \quad
     F_{IR_i}=\varphi_i{\cdots}(\varphi_2(\varphi_1(I_{IR}))),   
\end{equation}
where $\phi_i$ and $\varphi_i$ represent the convolutional blocks of RGB and IR branches at the $i$-th stage, respectively.

To implement cross-modality feature fusion, existing methods~\cite{guan2019fusion,qingyun2021cross,shen2024icafusion} primarily emphasize the integration of spatial features, yet they inadequately consider the feature disparity between modalities.
Consequently, the fused model fails to effectively model the correlations between targets of different modalities, diminishing the model's representation capacity.
Motivated by Mamba~\cite{gu2023mamba} with strong sequence modeling ability on the state space, we design a Fusion-Mamba block~(FMB) to construct a hidden state space for cross-modal feature interaction and association. The effectiveness of FMB lies in two key modules, the State Space Channel Swapping~(SSCS) module and the Dual State Space Fusion~(DSSF) module, which can reduce disparities between cross-modality features to enhance the representation consistency of fused features. Alg.~\ref{alg:AOA} provides the computation process of SSCS and DSSF modules.

\textbf{SSCS module.} This module aims to enhance cross-modality feature interaction for shallow feature fusion through the channel swapping operation and VSS block. Cross-modal feature correlation is constructed by integrating information from distinct channels, which enriches the diversity of channel features to improve fusion performance. First, we employ the channel swapping operation to generate new local features of RGB $T_{R_i}$ and IR $T_{IR_i}$, which can be formulated as:
\begin{equation}
T_{R_i}=CS(F_{R_i},F_{IR_i}), \quad T_{IR_i}=CS(F_{IR_i},F_{R_i}),
\end{equation}
where $CS(\cdot,\cdot)$ is the channel swapping operation, which is easy to implement by channel splitting and concatenation. First, both local features $F_{R_i}$ and $F_{IR_i}$ are divided into four equal parts along the channel dimension. Subsequently, we select the first and third parts from $F_{R_i}$, and the second and fourth parts from $F_{IR_i}$ through concatenation in a portion order to generate new local RGB features $T_{R_i}$. Correspondingly, we generate new local IR features $T_{IR_i}$. After that, one VSS block is applied to $T_{R_i}$ and $T_{IR_i}$, which enhances cross-modality interaction from shallow features:
%
\begin{equation}
\tilde{F}_{R_i}=VSS(T_{R_i}), \quad \tilde{F}_{IR_i}=VSS(T_{IR_i}),
\end{equation}
where $VSS(\cdot)$ denotes the VSS block~\cite{liu2024vmamba} depicted in Fig.~\ref{fig:Architecture}.
$\tilde{F}_{R_i}$ and $\tilde{F}_{IR_i}$ are the outputs of shallow fused features from RGB and IR modality, respectively.


\textbf{DSSF module.}
To further reduce modality disparities, we build a hidden state space for cross-modality feature association and complementary.
DSSF is proposed to model cross-modality object correlation to facilitate feature fusion.
%
%
Specifically, we employ the VSS block by projecting features from both modalities into a hidden state space, and a gating mechanism is utilized to construct hidden state transitions dually for cross-modality deep feature fusion.

Formally, after obtaining the shallow fused features $\tilde{F}_{R_i}$ and $\tilde{F}_{IR_i}$, we first project them into the hidden state space through a VSS block without gating as:
\begin{equation}
    y_{R_i}=P_{in}(\tilde{F}_{R_i}), \quad y_{IR_i}=P_{in}(\tilde{F}_{IR_i})
\label{eq8}
\end{equation}
where $P_{in}(\cdot)$ denotes an operation for projecting features to the hidden state space. The detailed implementation is described in lines 13-17 of Alg.~\ref{alg:AOA}. $y_{R_i}$ and $y_{IR_i}$ denote the hidden state features. 
We also project $\tilde{F}_{R_i}$ and $\tilde{F}_{IR_i}$ to obtain gating parameters $z_{R_i}$ and $z_{IR_i}$:
\begin{equation}
    z_{R_i}=f_{\theta_i}(\tilde{F}_{R_i}),\quad z_{IR_i}=g_{\omega_i}(\tilde{F}_{IR_i})
\label{eq9}
\end{equation}
where $f_{\theta_i}(\cdot)$ and $g_{\omega_i}(\cdot)$ represent the gating operation with parameters $\theta_i$ and $\omega_i$ in a dual stream, respectively.
After that, we employ the gating outputs of $z_{R_i}$ and $z_{IR_i}$ in Eq.~\ref{eq9} to modulate $y_{R_i}$ and $y_{IR_i}$, and implement the hidden state feature fusion as: 
\begin{equation}
 y'_{R_i}=y_{R_i} \cdot z_{R_i} + z_{R_i} \cdot y_{IR_i},  
\label{eq10}       
\end{equation}
\begin{equation}
y'_{IR_i}=y_{IR_i} \cdot z_{IR_i} + z_{IR_i} \cdot y_{R_i},
\label{eq10_2}       
\end{equation}
where $y'_{R_i}$ and $y'_{IR_i}$ represent the hidden state feature of RGB and IR after feature interaction, respectively. $\cdot$ is an element-wise product. Actually, Eq.~\ref{eq10} and ~\ref{eq10_2} build the cross-modality fusion in a hidden state space based on the gating mechanism, and dual attention is fully used for cross-branch information complementarity.

Subsequently, we project $y'_{R_i}$ and $y'_{IR_i}$ back to the original space and pass them through a residual connection to obtain the complementary features $\overline{F}_{R_i}$ and $\overline{F}_{IR_i}$:
\begin{equation}
    \overline{F}_{R_i} = P_{out}(y'_{R_i}) + \tilde{F}_{R_i}, \quad
    \overline{F}_{IR_i} = P_{out}(y'_{IR_i}) + \tilde{F}_{IR_i}.
\label{eq11}
\end{equation}
where $P_{\text{out}}(\cdot)$ denotes the projection operation with a linear transformation. 

In practice, we stack several DSSF modules~(\emph{i.e.}, Eq.~\ref{eq8} to Eq.~\ref{eq11}) to obtain much deeper feature fusion, which achieves better results. 
However, the number of DSSF modules will saturate at a certain value, which is further evaluated in our experiments.  
Finally, we merge the complementary features into the local features to enhance feature representation by the addition operation:
\begin{equation}
    \hat{F}_{R_i} = F_{R_i} + \overline{F}_{R_i}, \quad
    \hat{F}_{IR_i} = F_{IR_i} + \overline{F}_{IR_i}.
\label{eq12}
\end{equation}

\begin{algorithm}[t]
    \caption{Algorithm of Fusion-Mamba block~(FMB)} 
    \label{alg:AOA}
    \footnotesize
    \renewcommand{\algorithmicrequire}{\textbf{Input:}}
    \renewcommand{\algorithmicensure}{\textbf{Output:}}
    \begin{algorithmic}[1]
        \REQUIRE $F_{R_i}: \textcolor{green!80!black}{(B,C_i,H_i,W_i)}$, $F_{IR_i}: \textcolor{green!80!black}{(B,C_i,H_i,W_i)}$  
        \ENSURE $\hat{F}_{R_i}: \textcolor{green!80!black}{(B,C_i,H_i,W_i)}$, $\hat{F}_{IR_i}: \textcolor{green!80!black}{(B,C_i,H_i,W_i)}$    

        \STATE $\textcolor{red!80!black}{/* ~State ~Space ~Channel ~Swapping ~(SSCS) ~module~*/}$
        \STATE $\textcolor{gray!100!black}{/* ~channel ~swapping ~with ~shallow ~feature ~fusion ~*/}$
        \STATE $T_{R_i}: \textcolor{green!80!black}{(B,C_i,H_i,W_i)} \leftarrow CS(F_{R_i},F_{IR_i})$
        \STATE $T_{IR_i}: \textcolor{green!80!black}{(B,C_i,H_i,W_i)} \leftarrow CS(F_{IR_i},F_{R_i})$
        \STATE $\tilde{F}_{R_i}: \textcolor{green!80!black}{(B,C_i,H_i,W_i)} \leftarrow VSS(T_{R_i})$
        \STATE $\tilde{F}_{IR_i}: \textcolor{green!80!black}{(B,C_i,H_i,W_i)} \leftarrow VSS(T_{IR_i})$
        \STATE $\textcolor{red!80!black}{/* ~Dual ~State ~Space ~Fusion ~(DSSF) ~module~*/}$
        \FOR{k = 1 to N} 
        \IF{k != 1}
        \STATE $\tilde{F}_{R_i} \leftarrow \overline{F}_{R_i}$
        \STATE $\tilde{F}_{IR_i} \leftarrow \overline{F}_{IR_i}$
        \ENDIF
            \FOR{i in \{$R_i$, $T_i$\}}
                \STATE $\textcolor{gray!100!black}{/* ~Project ~the ~input ~into ~the ~hidden ~state ~space ~*/}$
                \STATE $x_i: \textcolor{green!80!black}{(B,P_i,H_i,W_i)} \leftarrow Linear(Norm(\tilde{F}_i))$
                \STATE $x'_i: \textcolor{green!80!black}{(B,P_i,H_i,W_i)} \leftarrow SiLU(DWConv(x_i))$
                \STATE $y_i: \textcolor{green!80!black}{(B,P_i,H_i,W_i)} \leftarrow Norm(SS2D(x'_i))$
            \ENDFOR
            \STATE $\textcolor{gray!100!black}{/* ~Obtain ~the ~gating ~outputs ~z_{R_i}, ~z_{IR_i} ~*/}$
            \STATE $z_{R_i}: \textcolor{green!80!black}{(B,P_i,H_i,W_i)} \leftarrow SiLU(Linear(Norm(\tilde{F}_{R_i}))$
            \STATE $z_{IR_i}: \textcolor{green!80!black}{(B,P_i,H_i,W_i)} \leftarrow SiLU(Linear(Norm(\tilde{F}_{IR_i}))$
            \STATE $\textcolor{gray!100!black}{/* ~Transition ~of ~hidden ~states ~*/}$
            \STATE $y'_{R_i}: \textcolor{green!80!black}{(B,P_i,H_i,W_i)} \leftarrow y_{R_i} \bullet SiLU(z_{R_i})+ SiLU(z_{R_i}) \bullet y_{IR_i}$
            \STATE $y'_{IR_i}: \textcolor{green!80!black}{(B,P_i,H_i,W_i)} \leftarrow y_{R_i} \bullet SiLU(z_{IR_i})+ SiLU(z_{IR_i})\bullet y_{IR_i}$
            \STATE $\textcolor{gray!100!black}{/* ~Project ~the ~hidden ~state ~into ~the ~original ~space ~*/}$
            \STATE $\overline{F}_{R_i}: \textcolor{green!80!black}{(B,C_i,H_i,W_i)} \leftarrow Linear(y'_{R_i})+\tilde{F}_{R_i}$
            \STATE $\overline{F}_{IR_i}: \textcolor{green!80!black}{(B,C_i,H_i,W_i)} \leftarrow Linear(y'_{IR_i})+\tilde{F}_{IR_i}$
        \ENDFOR
        \STATE $\textcolor{gray!100!black}{/* ~Enhance ~feature ~representation ~*/}$
        \STATE $\hat{F}_{R_i}: \textcolor{green!80!black}{(B,C_i,H_i,W_i)} \leftarrow F_{R_i}+\overline{F}_{R_i}$
        \STATE $\hat{F}_{IR_i}: \textcolor{green!80!black}{(B,C_i,H_i,W_i)} \leftarrow F_{IR_i}+\overline{F}_{IR_i}$
        \RETURN $\hat{F}_{R_i}$,$\hat{F}_{IR_i}$
    \end{algorithmic}
\end{algorithm}

\subsubsection{Loss Function}
\label{sec:Loss}
After FMB, the enhanced features from RGB and IR~(\emph{i.e.}, $\hat{F}_{R_i}$  and $\hat{F}_{IR_i}$ in Eq.~\ref{eq12}) are further added to generate fused feature $P_i$ as the input of the neck to improve the detection performance. Following~\cite{jocher2022ultralytics,Jocher_Ultralytics_YOLO_2023}, the total loss function can be constructed as:
\begin{equation}
    \mathcal{L} = \lambda_{\text{coord}} \mathcal{L}_{\text{coord}} + \mathcal{L}_{\text{conf}} + \mathcal{L}_{\text{class}},
\label{loss}
\end{equation}
where $\lambda_{\text{coord}}$ is a hyperparameter that adjusts the weight of the localization loss $\mathcal{L}_{\text{coord}}$, $\mathcal{L}_{\text{conf}}$ is the confidence loss, and $\mathcal{L}_{\text{class}}$ is the classification loss. 
The more details of individual loss term for $\mathcal{L}_{\text{coord}}, \mathcal{L}_{\text{conf}}$ and $\mathcal{L}_{\text{class}}$ are described in~{jocher2022ultralytics}.




\begin{table*}[t]
	\centering
	\caption{Comparison results with SOTA methods on LLVIP dataset. The  best and second results are highlighted in red and green, respectively.}
 \vspace{-1em}
	\setlength{\tabcolsep}{4.5mm}
        \resizebox{0.7\textwidth}{!}{  
	\begin{tabular}{lcc|cc}
		\toprule 
		Methods & Modality & Backbone & $m$AP$_{50}$ & $m$AP \\
		
		\midrule
		(NIPS'16) Faster R-CNN&IR&ResNet50&92.6&50.7\\
		(CVPR'18) Cascade R-CNN&IR&ResNet50&95.0&56.8\\
        
           ~(CVPR'23) DDQ DETR&IR&ResNet50&93.9&58.6\\
           ~(Zenodo'22) YOLOv5-l&IR&YOLOv5&94.6&61.9\\
           ~(2023) YOLOv8-l&IR&YOLOv8&95.2&62.1\\
             Faster R-CNN~\cite{ren2015faster}&RGB&ResNet50&88.8&47.5\\
		Cascade R-CNN~\cite{cai2018cascade}&RGB&ResNet50&88.3&47.0\\
		
		DDQ DETR~\cite{zhang2023dense}&RGB&ResNet50&86.1&46.7\\
            YOLOv5-l~\cite{jocher2022ultralytics}&RGB&YOLOv5&90.8&50.0\\
		YOLOv8-l~\cite{Jocher_Ultralytics_YOLO_2023}&RGB&YOLOv8&91.9&54.0\\
		(2016) Halfway fusion~\cite{liu2016multispectral}&IR+RGB&VGG16&91.4&55.1\\
		(WACV'21) GAFF~\cite{zhang2021guided}&IR+RGB&Resnet18&94.0&55.8\\
		(ECCV'22) ProEn~\cite{chen2022multimodal}&IR+RGB&ResNet50&93.4&51.5\\
		(CVPR'23) CSAA~\cite{cao2023multimodal}&IR+RGB&ResNet50&94.3&59.2\\
           ~(2024) RSDet~\cite{zhao2024removal}&IR+RGB&ResNet50&95.8&61.3\\
		(IF'23) DIVFusion~\cite{tang2023divfusion}&IR+RGB&YOLOv5&89.8&52.0\\
		\textbf{Fusion-Mamba(Ours)}&IR+RGB&YOLOv5&\textcolor{green}{96.8}&\textcolor{green}{62.8}\\
		\textbf{Fusion-Mamba(Ours)} &IR+RGB&YOLOv8&\textcolor{red}{97.0}&\textcolor{red}{64.3}\\
		\bottomrule
	\end{tabular}
        }
	\label{tab:LLVIP}
\end{table*}


\subsection{Compared to Transformer-based fusion}
\label{sec:Compared}

Existing Transformer-based cross-modality fusion methods~\cite{qingyun2021cross,shen2024icafusion} flatten and concatenate the features with convolution to generate the intermediate fused features, which is further fused by multi-head cross-attention to generate final fused features.
They cannot effectively reduce the modality disparities just on the spatial interaction, which is due to the difficult modeling of object correlation from cross-modality features.
Our FMB block can scan features in four directions to obtain four sets of patches and effectively preserve the local information of the features.
In addition, these patches are mapped into a hidden space for feature fusion.
This mapping-based deep feature fusion method effectively reduces spatial disparities through dual-direction gated attention, which further suppresses redundant features and captures complementary information among modalities. As such, the proposed FMB reduces disparities between cross-
modality features and enhances the representation consistency of
fused features. 

In addition, the time complexity of the Transformer's global attention is $O(N^2)$, while Mamba's time complexity is only $O(N)$, where $N$ is the sequence length.
From the experiment perspective, using the same detection model architecture, replacing the transformer-based fusion module with Fusion-Mamba block saves $7-19$ms inference time on one paired images. More details are discussed in our experiments.

\section{Experiments}
\subsection{Experimental Setups}

\textbf{Datasets.} Our Fusion-Mamba method is evaluated on three widely-used visible-infrared benchmark datasets, LLVIP~\cite{jia2021llvip}, $M^3FD$~\cite{liu2022target} and FLIR~\cite{reference1}.

LLVIP is an aligned visible and infrared~(IR) dataset collected in low-light environments for pedestrian detection, which contains $15,488$ RGB-IR image pairs. Following the official standards, we use $12,025$ pairs for training and $3,463$ pairs for testing.

$M^3FD$ contains $4,200$ RGB and IR-aligned image pairs collected in various environments including different lighting, seasons, and weather scenarios.
It has six categories usually appearing in autonomous driving and road monitoring. Since there is no official dataset partitioning method, we use train/test splits provided by~\cite{liang2023explicit}.

FLIR is collected in day and night scenes with five categories: \textit{people}, \textit{car}, \textit{bike}, \textit{dog}, and \textit{other cars}.
Following by~\cite{zhang2020multispectral}, we use the FLIR-Aligned with $4,129$ pairs for training and $1,013$ for testing.

\textbf{Evaluation metrics.}
We use the most common evaluation metric $m$AP and $m$AP$_{50}$. 
The $m$AP$_{50}$ metric represents the mean AP under IoU $0.50$ and The $m$AP metric represents the mean AP under IoU ranges from $0.50$ to $0.95$ with a stride of $0.05$~\cite{zhao2024removal}.
The larger test value of the two metrics means better model performance. We also report the average inference time of our method evaluated on one A800 GPU with $5$ runs on the input size of $640\times640$.

\textbf{Implementation details.}
All the experiments are implemented in the two-stream framework~\cite{qingyun2021cross} with a single GPU A800.
The backbone, neck and head structures of our Fusion-Mamba are default set the same as those in YOLOv5-l or YOLOv8-l.
During training, we set the batch size to $4$,
SGD optimizer is set with a momentum of $0.9$ and a weight decay of $0.001$.
The
input image size is $640 \times 640$ for all three datasets and the training epoch is set to $150$ with an initial learning rate of $0.01$. The number of SSCS and DSSF modules in FMB is default set to $1$ and $8$, respectively. $\lambda_{\text{coord}}$ is set to $7.5$. Other training hyper-parameters are the same as YOLOv8.

\begin{table*}[t]
	\begin{center}
		\caption{Comparison results with eight SOTA methods on $M^3FD$ Dataset. The best results are highlighted in bold.}
  \vspace{-1em}
		\setlength{\tabcolsep}{2mm}{
			\begin{tabular}{lccc|ccccccc}
				\toprule
				Methods&Backbone&$m$AP$_{50}$&$m$AP&People&Bus&Car&Motorcycle&Lamp&Truck\\
				\midrule
				
				(2020) DIDFuse~\cite{zhao2020didfuse}&YOLOv5&78.9&52.6&79.6&79.6&92.5&68.7&84.7&68.7\\
				(IJCV'21) SDNet~\cite{zhang2021sdnet}&YOLOv5&79.0&52.9&79.4&81.4&92.3&67.4&84.1&69.3\\
				(CVPR'22) RFNet~\cite{xu2022rfnet}&YOLOv5&79.4&53.2&79.4&78.2&91.1&72.8&85.0&69.0\\
				(CVPR'22) TarDAL~\cite{liu2022target}&YOLOv5&80.5&54.1&81.5&81.3&\textbf{94.8}&69.3&87.1&68.7\\
				(MM'22) DeFusion~\cite{sun2022detfusion}&YOLOv5&80.8&53.8&80.8&83.0&92.5&69.4&\textbf{87.8}&71.4\\
               ~(CVPR'23) CDDFuse~\cite{zhao2023cddfuse}&YOLOv5&81.1&54.3&81.6&82.6&92.5&71.6&86.9&71.5\\
				(MM'23) IGNet~\cite{li2023learning}&YOLOv5&81.5&54.5&81.6&82.4&92.8&73.0&86.9&72.1\\
               ~(JAS'22) SuperFusion~\cite{tang2022superfusion}&YOLOv7&83.5&56.0&\textbf{83.7}&\textbf{93.2}&91.0&\textbf{77.4}&70.0&85.8\\
				\textbf{Fusion-Mamba~(ours)}&YOLOv5&\textbf{85.0}&\textbf{57.5}&80.3&92.8&91.9&73.0&84.8&\textbf{87.1}\\
                 \midrule
                YOLOv8l-IR&YOLOv8&79.5&53.1&82.9&90.9&90.0&64.6&63.0&85.9\\
               
				YOLOv8l-RGB~\cite{Jocher_Ultralytics_YOLO_2023}&YOLOv8&80.9&52.5&70.6&92.9&91.2&69.6&75.3&86.0\\
				
				\textbf{Fusion-Mamba~(ours)}&YOLOv8&\textbf{88.0}&\textbf{61.9}&\textbf{84.3}&\textbf{94.2}&\textbf{92.9}&\textbf{80.5}&\textbf{87.5}&\textbf{88.8}\\
				\bottomrule
			\end{tabular}
			\label{tab:M3FD}
		}
	\end{center}
\end{table*}

\subsection{Comparison with SOTA Methods}
To verify the effectiveness of our Fusion-Mamba method, we employ two backbones based on YOLOv5 and YOLOv8 to make a fair comparison with SOTA methods.\\ 
\textbf{LLVIP Dataset.}
The results of different methods on LLVIP are summarized in Tab.~\ref{tab:LLVIP}.
We compare the proposed Fusion-Mamba method using two different backbones with $6$ SOTA multi-spectral object detection methods and $5$ single-modality detection methods.
For single-modality detection, the detection performance only using IR images is better than that only using RGB images, which is due to the effect of low light conditions.
After feature fusion with RGB and IR, the $m$AP performance is improved based on ResNet backbones, outperforming the single IR modality detection.
For example, RSDet with the ResNet50 backbone outperforms Cascade R-CNN only using IR modality by $4.5\%$ $m$AP. 
Note that it fails to effective fusion on the YOLOv5 backbone, \emph{e.g.}, a simple YOLOv5 detection framework using IR modality input achieves $61.9\%$ $m$AP, significantly outperforming fusion method DIVFusion by $9.9\%$ $m$AP. With the same YOLOV5 backbone, 
our Fusion-Mamba method achieves $m$AP gain over the YOLOv5 detection framework only with IR by $0.9\%$, and also outperforms the best previous fusion method RSDet by $1.5\%$ $m$AP. To explain, our SSCS and DSSF effectively reduce modality disparities to improve the representation consistency of fused features. 
our method is also effective for the YOLOv8 backbone, which achieves state-of-the-art performance with $97.0\%$ $m$AP$_{50}$ and $64.3\%$ $m$AP.

\textbf{\textbf{$M^3FD$} Dataset.}
We compare our method with $7$ SOTA detectors based on the YOLOv5 and $1$ SOTA detector based on YOLOv7.
As shown in Tab.~\ref{tab:M3FD}, our Fusion-Mamba performs best on the 
 all categories with $m$AP$_{50}$ and $m$AP metric compared with SOTA methods based on the same YOLOv5 backbone and our method based on YOLOv8 backbone achieves new SOTA results on the \textit{People}, \textit{Bus} \textit{Motorcycle} and \textit{Truck} categories, 
 while $m$AP$_{50}$ and $m$AP metric further increases by $3\%$ and $4.4\%$.
In addition, our method using YOLOv5 backbone also outperforms SuperFusion based on YOLOv7 by $1.5\%$ $m$AP and $m$AP$_{50}$, despite the lower feature representation ability of YOLOv5 than YOLOv7. This is due to the effectiveness of our FMB, improving the inherent complementary of cross-modality features.

\begin{table}[t]
    \centering
    \caption{Comparison results with SOTA methods on FLIR-Aligned Dataset. The best results are highlighted in bold.}
    \vspace{-1em}
    \begin{adjustbox}{width=0.5\textwidth}
    \begin{tabular}{lccccc}
        \toprule
        Methods & Backbone & $m$AP$_{50}$ & $m$AP & Param. & Time(ms) \\
        \midrule
        (2021)CFT~\cite{qingyun2021cross} & YOLOv5 & 78.7 & 40.2 & \textbf{206.0M} & 68 \\
        (PRL'24)CrossFormer\cite{lee2024crossformer} & YOLOv5 & 79.3 & 42.1 & 340.0M & 80 \\
        ~(2024)RSDet~\cite{zhao2024removal} & ResNet50 & 81.1 & 41.4 & - & - \\
        \textbf{Fusion-Mamba~(ours)} & YOLOv5 & \textbf{84.3} & \textbf{44.4} & 244.6M & \textbf{61} \\
        \cline{1-6}
        YOLOv8l-IR & YOLOv8 & 72.9 & 38.3 & 76.7M & 22 \\
        YOLOv8l-RGB & YOLOv8 & 66.3 & 28.2 & 76.7M & 22 \\
        \textbf{Fusion-Mamba~(ours)} & YOLOv8 & \textbf{84.9} & \textbf{47.0} & 287.6M & 78 \\
        \bottomrule
    \end{tabular}
    \end{adjustbox}
    \label{tab:FLIR}
\end{table}

\begin{figure}[t]
    \centering 
    \includegraphics[width=0.92\linewidth]{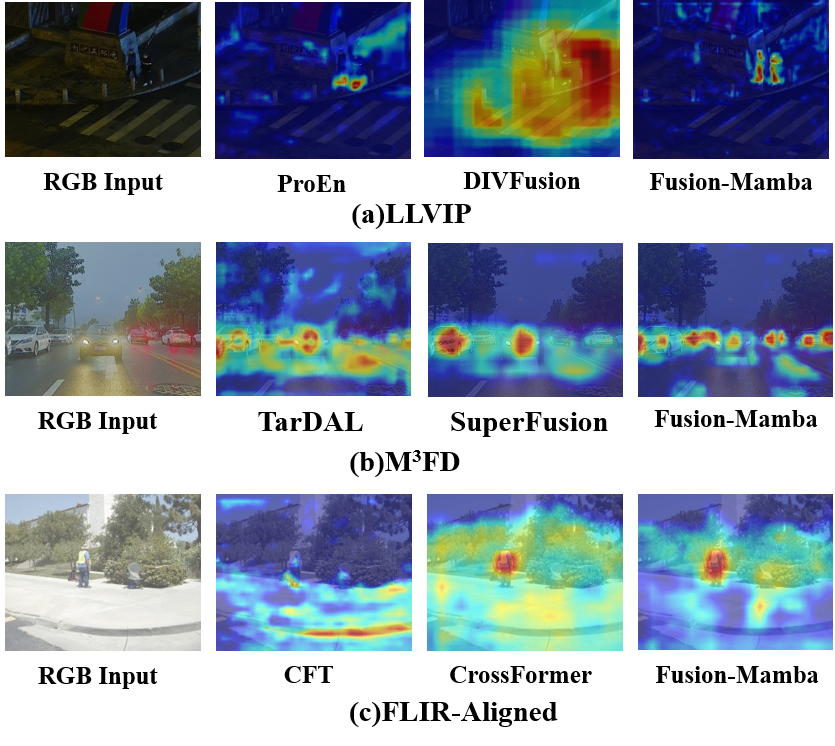} 
    \vspace{-1em}
    \caption{Heatmap visualization  of various cross-modality object detection methods on LLVIP, $M^3$FD  and FLIR-Aligned datasets.}
    \label{fig:heatmapex}
\end{figure}

\textbf{FLIR-Aligned Dataset.} 
As shown in Tab.~\ref{tab:FLIR}, Fusion-Mamba also performs best on Aligned-FLIR Dataset.
Compared to CrossFormer based on the two-stream YOLOv5 Backbone, our method based on YOLOv8 and YOLOv5 surpasses them with $5.6\%$ and $5\%$ on $m$AP$_{50}$, and $4.9\%$ and $2.3\%$ on $m$AP, respectively.
We also outperform RSDet with $3.8\%$ $m$AP$_{50}$ and $5.6\%$ $m$AP.
In terms of speed, our method with YOLOv5 achieves the fastest speed, saving $7$ms and $19$ms for one paired images detection, compared to transformer-based CFT and CrossFormer methods. For parameters, our method based on YOLOv5 saves about $100$M compared to the CrossFormer method. 
Despite our method based on YOLOv8 increases about $40$M parameters than that on YOLOv5, the $m$AP is significantly increased by $2.6\%$.
This result indicates our method based on hidden space modeling better integrates features between different modalities, suppressing modality disparities to enhance the representation ability of fused features with the best trade-off between performance and computation cost.\par
\textbf{Visialization of heatmaps.}
To visually demonstrate the high performance of our model, we randomly select one paired images from each of the three experimental datasets to visualize $P_5$ heatmap, and compare our method with other fusion methods. As shown in Fig.\ref{fig:heatmapex}, compared to other methods, our model focuses more on the targets rather than dispersing or focusing on unrelated parts. More examples are presented in the supplementary materials. We also visualize the object detection results to evaluate the effectiveness of our method in the supplementary materials.

\subsection{Ablation Study}
We use the FLIR-Aligned dataset for ablation study to separately verify the effectiveness of SSCS and DSSF modules and further explore the influence of numbers and position of the DSSF module. In particular, we also evaluate the effect of dual attention of DSSF module.
All of these experiments are conducted based on YOLOv8 backbone.

\textbf{Effects of SSCS and DSSF modules}
The results of removing SSCS and DSSF in the FMB are summarized in Tab.~\ref{tab:FLIR_ablation}. After removing SSCS module~(second row in Tab.~\ref{tab:FLIR_ablation}), the detector performance is decreased by $2\%$ and $1.1\%$ on $m$AP$_{50}$ and $m$AP, respectively.
To explain, without the initial exchange of two modal features and shallow mapping fusion, the feature disparity is not well reduced during the following deep fusion.
Meanwhile, without DSSF~(third row in Tab.~\ref{tab:FLIR_ablation}), only shallow fusion interaction could not effectively suppress redundant features and activate effective features during feature fusion, leading to a decrease in detector performance with $2.5\%$ and $2.4\%$ dropping on $m$AP$_{50}$ and $m$AP, respectively. 
Both SSCS and DSSF are removed and the fused features are directly obtained by the addition of two local modality features~(fourth row in Tab.~\ref{tab:FLIR_ablation}), whose performance is significantly reduced by $4.8\%$ and $7.6\%$ on $m$AP$_{50}$ and $m$AP, respectively. 
These results demonstrate that these two components of FMB are effective for cross-modality object detection.

\begin{table}
\centering
\caption{Effects of SSCS and DSSF on FLIR-Aligned Dataset.}
\vspace{-1em}
\resizebox{0.45\textwidth}{!}{%
\begin{tabular}{lccccc}
\toprule
Methods & $m$AP$_{50}$ & $m$AP$_{75}$ & $m$AP & Param. & Time(ms) \\
\midrule
Fusion-mamba & 84.9 & 45.9 & 47.0 & 287.6M & 78 \\
Removing SSCS & 82.9 & 42.3&45.9&266.8M&69\\
Removing DSSF & 82.4 & 42.0&44.6&138.0M&57\\
Removing SSCS \& DSSF & 80.1 & 36.3 & 39.4 & 117.2M&48\\
\bottomrule
\end{tabular}%
}
\label{tab:FLIR_ablation}
\end{table}

\begin{table}
  \begin{center}
  \caption{Effect of FMB position on FLIR-Aligned Dataset.}
  \vspace{-1em}
  \setlength{\tabcolsep}{1mm}{
  \resizebox{0.4\textwidth}{!}{
  \begin{tabular}{lccccc}
    \toprule
    Positions&$m$AP$_{50}$&$m$AP$_{75}$&$m$AP&Param.&Time(ms)\\
    \midrule
    $\{P_2, P_3, P_5\}$&83.9&44.4&46.7&256.8M&72\\
    $\{P_2,P_4, P_5\}$&76.1&39.8&42.3&281.1M&75\\
   $\{P_3, P_4, P_5\}$&84.9&45.9&47.0&287.6M&78\\
  \bottomrule
\end{tabular}
\label{tab:FLIR_ablation4}
}
}
\end{center}
\end{table}

\textbf{Effect on the position of FMB.} Following the work~\cite{qingyun2021cross,lee2024crossformer}, we also set three FMB for feature fusion. Here, we will further explore the impact of the FMB position, which stages should add FMB.
We select three groups of multi-level features: $\{P_2, P_3, P_5\}$, $\{P_2,P_4, P_5\}$ and $\{P_3, P_4, P_5\}$ for ablation study, where $P_i$ is the fused feature at the $i$-th stage using FMB.
As presented in Tab.~\ref{tab:FLIR_ablation4}, the position $\{P_3, P_4, P_5\}$ achieves the best trade-off between performance and computation complexity. Thus, we default select this position for experiments.

\textbf{Effect on the numbers of DSSF modules.}
We have validated the effectiveness of DSSF in Tab.~\ref{tab:FLIR_ablation}. Here, we further evaluate the effect of the number of DSSF modules, as summarized in Tab.~\ref{tab:FLIR_ablation2}. We
select four kinds of DSSF numbers~(\textit{i.e.}, $2$, $4$, $8$, $16$), and keep other model settings be consistent with the above experiments.
We can see that the number of blocks is set to $8$, which achieves the best performance. 
The $8$ DSSF modules will be saturated, increasing this number causes the drift of complementary features, which leads to a decrease in fusion performance.
\begin{table}
  \begin{center}
  \caption{Effect on the numbers of DSSF modules on FLIR-Aligned Dataset.}
  \vspace{-1em}
  \setlength{\tabcolsep}{1mm}{
  \resizebox{0.4\textwidth}{!}{
  \begin{tabular}{lccccc}
    \toprule
    Number&$m$AP$_{50}$&$m$AP$_{75}$&$m$AP&Param.&Time(ms)\\
    \midrule
    2&82.3&42.8&45.5&175.3M&50\\
    4&82.6&43.6&45.9&212.7M&65\\
    8&84.9&45.9&47.0&287.6M&78\\
    16&83.4&44.2&46.3&437.2M&148\\
  \bottomrule
\end{tabular}
\label{tab:FLIR_ablation2}
}
}
\end{center}
\end{table}


\textbf{Effect on the dual attention of DSSF module.}
To further explore the effectiveness of our gating mechanism whether using the dual attention of DSSF module, we separately remove the IR attention~(\emph{i.e.}, $z_{IR_i} \cdot y_{R_i}$ in Eq.~\ref{eq10}) in the RGB branch, the RGB attention~(\emph{i.e.}, $z_{IR_i} \cdot y_{R_i}$ in Eq.~\ref{eq10_2}) in the IR branch, and both dual attention. The results are shown in Tab.\ref{tab:FLIR_ablation3}. 
After removing IR attention or RGB attention, $m$AP$_{50}$ decreases $1.6\%$ or $1.1\%$ for reducing attention interaction between two features, respectively. When both dual attention is removed, the DSSF module becomes a stack of VSS blocks, and $m$AP$_{50}$ degrades $2\%$. It is noted that both IR and RGB attention branches share weights with the other branches, which is equivalent to only adding activation functions and feature addition operations, compared to the removal of dual attention. Thus, the usage of dual attention does not have a significant impact on model parameters and runtime, while it significantly improves the detection performance.
\begin{table}
  \begin{center}
  \caption{Ablation study of dual attention in the RGB and IR branches of DSSF module on FLIR-Aligned Dataset.}
  \vspace{-1em}
  \setlength{\tabcolsep}{1mm}{
  \resizebox{0.45\textwidth}{!}{
  \begin{tabular}{lccccc}
    \toprule
    Methods&$m$AP$_{50}$&$m$AP$_{75}$&$m$AP&Param.&Time(ms)\\
    \midrule
    Fusion-Mamba&84.9&45.9&47.0&287.6M&78\\
    Removing $z_{IR_i} \cdot y_{R_i}$ in Eq.~\ref{eq10}&83.3&42.8&45.3&287.6M&77\\
    Removing $z_{IR_i} \cdot y_{R_i}$ in Eq.~\ref{eq10_2} &83.8&43.9&46.2&287.6M&77\\
    Removing both dual attention&82.9&41.7&44.8&287.6M&76\\
  \bottomrule
\end{tabular}
\label{tab:FLIR_ablation3}
}
}
\end{center}
\end{table}
\section{Conclusion}
In this paper, we propose a novel Fusion-Mamba method with well-designed SSCS module and DSSF module for multi-modal feature fusion.
In particular, SSCS exchanges infrared and visible channel features for a  shallow feature fusion.
Subsequently, DSSF is further designed for deeper multi-modal feature interaction in a hidden state space based on Mamba, and  a gated attention is used to suppress redundant features to enhance the effectiveness of feature fusion.
Extensive experiments conducted on three public RGB-IR datasets demonstrated that our method achieves new state-of-the-art performance with a higher inference efficiency than Transformers.
Our work confirms the potential of Mamba for cross-modal fusion, and we believe that our work can inspire more research on the application of Mamba in cross-modal tasks.
{
    \small
   \bibliographystyle{ieeenat_fullname}
   \bibliography{main}
}

\clearpage
\setcounter{page}{1}
\maketitlesupplementary

\section{More Heatmap Visualization Results}
\begin{figure}[t]
    \centering 
    \includegraphics[width=0.8\linewidth]{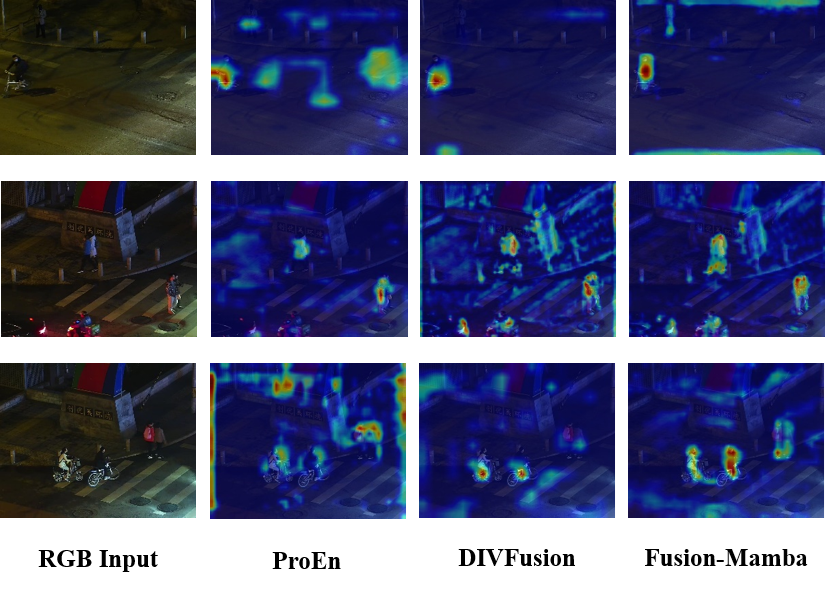} 
    \vspace{-1em}
    \caption{Comparison of heatmap visualization on LLVIP dataset.}
    \label{fig:supple1}
\end{figure}

\begin{figure}[t]
    \centering 
    \includegraphics[width=0.8\linewidth]{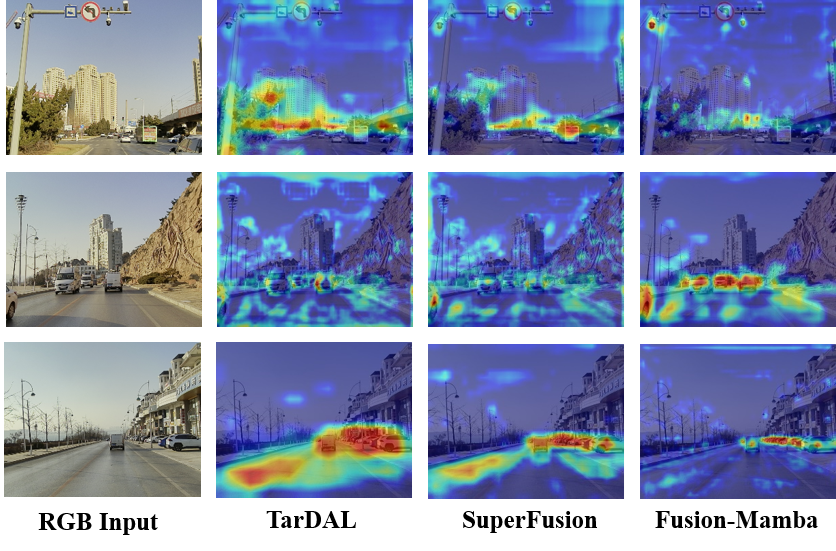} 
    \vspace{-1em}
    \caption{Comparison of heatmap visualization on $M^3FD$ dataset.}
    \label{fig:supple3}
\end{figure}

\begin{figure}[t]
    \centering 
    \includegraphics[width=0.8\linewidth]{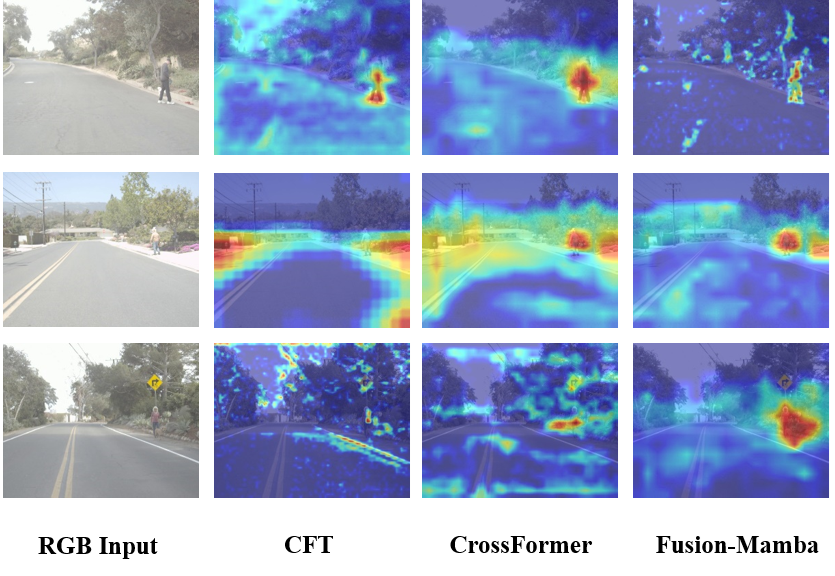} 
    \vspace{-1em}
    \caption{Comparison of heatmap visualization on FLIR-Aligned dataset.}
    \label{fig:supple2}
\end{figure}

We randomly select images from LLVIP, $M^3FD$ and FLIR-Aligned dataset and visualize $P_5$  heatmaps of different fusion methods. As shown in Fig.\ref{fig:supple1}, Fig.\ref{fig:supple3}, Fig.\ref{fig:supple2}, visualization examples demonstrate that our Fusion-Mamba method for cross-modality feature fusion on the hidden state space more focuses on the target, compared to CNN and Transformer based fusion methods. It also indicates that our method can effectively model the
correlations between targets of different modalities.

\section{Visualization of Object Detection}
\begin{figure*}[t]
    \centering 
    \includegraphics[width=0.9\linewidth]{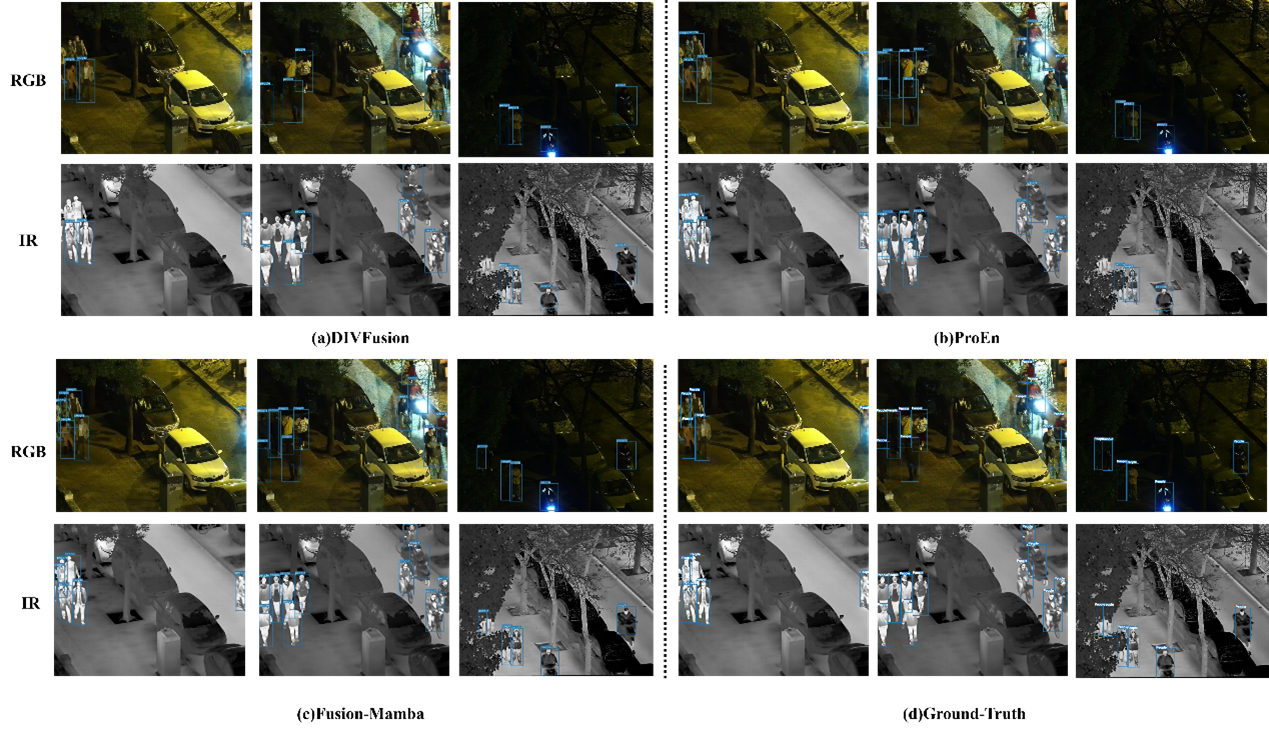} 
    \vspace{-1em}
    \caption{Detection results of three methods in LLVIP dataset.}
    \label{fig:10}
\end{figure*}

\begin{figure*}[t]
    \centering 
    \includegraphics[width=0.9\linewidth]{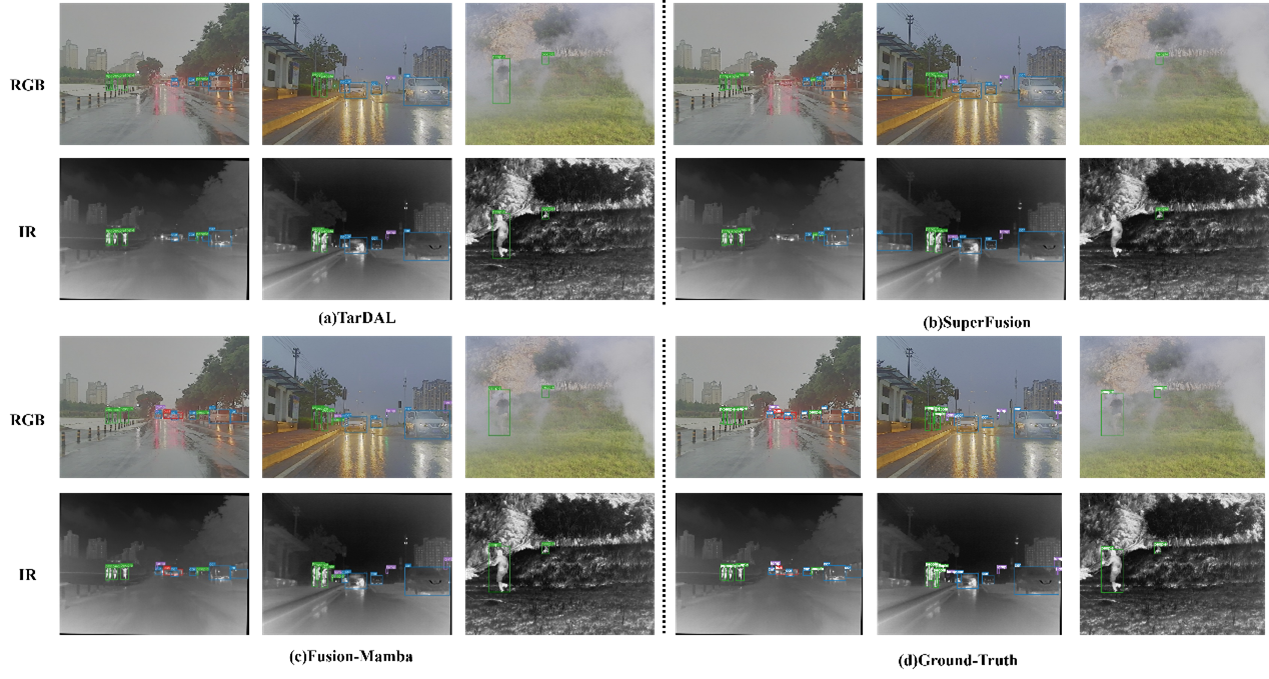} 
    \vspace{-1em}
    \caption{Detection results of three methods in $M^3FD$ dataset.}
    \label{fig:11}
\end{figure*}

\begin{figure*}[t]
    \centering 
    \includegraphics[width=0.9\linewidth]{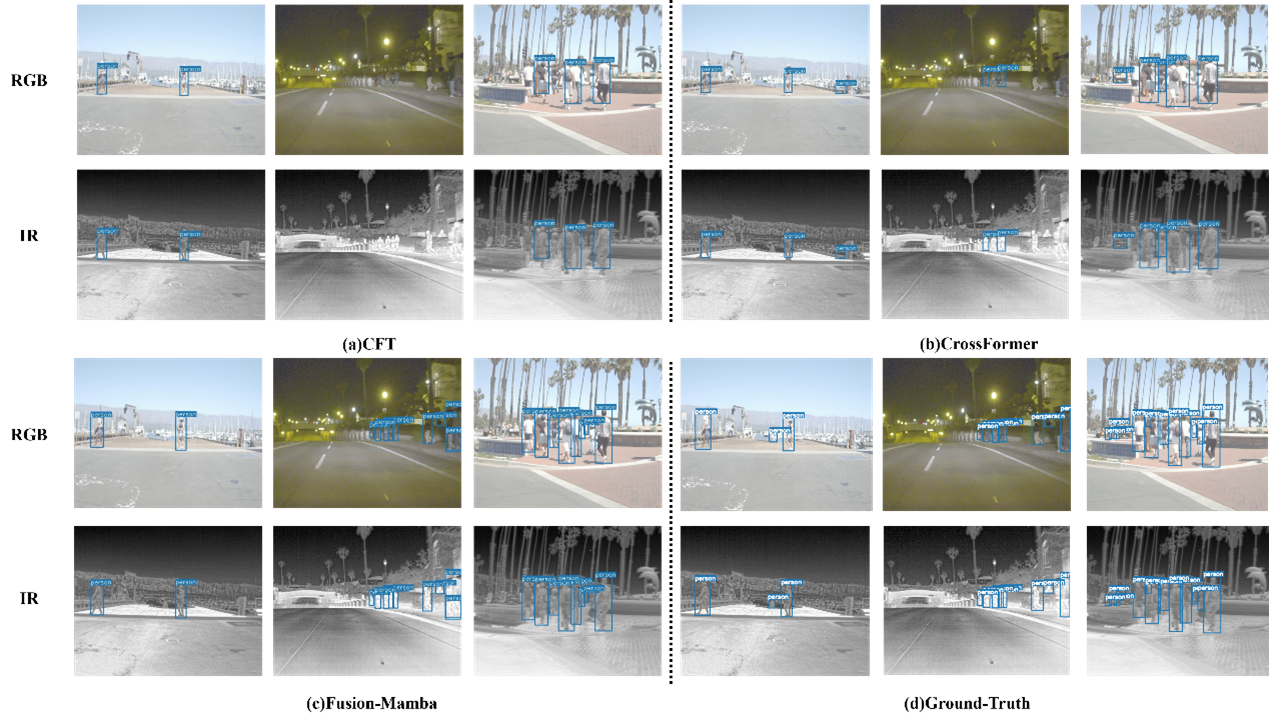} 
      \vspace{-1em}
    \caption{Detection results of three methods in FLIR-Aligned dataset.}
    \label{fig:12}
\end{figure*}

We also randomly select images from LLVIP, $M^3FD$ and FLIR-Aligned datasets and output bounding-box detection results using different fusion methods. 
As shown in Fig.\ref{fig:10}, Fig.\ref{fig:11}, Fig.\ref{fig:12}, compared to several SOTA methods, our method significantly reduces the missed detection results to improve mAP. 

For example, In Fig.\ref{fig:10} under the situations of insufficient lighting and severe occlusion, our method can detect more target objects compared to other methods, as our method's interaction based on hidden space effectively effectively integrates information from the IR modality.  
As shown in Fig.\ref{fig:11}, in harsh weather and situations where the target object is small and far away, our model can detect more diverse target objects compared to other methods, since our gated attention can effectively combine two modal information to suppress redundant features for better cross-modality object detection. 
In Fig.\ref{fig:12}, in target dense areas, our model can better distinguish and detect these targets, because our modal interaction method from shallow to deep can better preserve detailed information, while other SOTA methods tend to fail for the detection of dense objects.

Therefore, Our Fusion-Mamba method achieves the best detection performance, which improves
the perceptibility and robustness in complex scenes with several challenges, like low lighting, adverse weather, high occlusion, target density, and small target.

\end{document}